\documentclass[A4paper, 11 pt, conference]{ieeeconf}  
\IEEEoverridecommandlockouts       
\usepackage{geometry}
\usepackage{multirow}
\usepackage{amsmath}
\usepackage{booktabs}
\usepackage{multirow}
\usepackage{tabularx}
\usepackage{array}
\usepackage{xcolor}
\usepackage{graphics} % for pdf, bitmapped graphics files
\usepackage{epsfig} % for postscript graphics files
\usepackage{mathptmx} % assumes new font selection scheme installed
\usepackage{times} % assumes new font selection scheme installed
\usepackage{amsmath} % assumes amsmath package installed
\usepackage{amssymb}  % assumes amsmath package installed
\usepackage{subfigure}
\usepackage{algorithm} %format of the algorithm
\usepackage{algorithmic} %format of the algorithm
\usepackage{multirow} %multirow for format of table
\usepackage{hyperref}
\hypersetup{hidelinks}
\begin{document}
% Change to your title
\title{\LARGE \bf
Fast Implicit Neural Light Field Representation via Geometric Decomposition and Multi-Resolution Low-Rank Features}

\author{Yao Guo, Ligen Shi, Shuchen Sun, Jun Qiu and  Chang Liu$^*$ % <-this % stops a space
\thanks{
Yao Guo, Shuchen Sun, Jun Qiu and Chang Liu are with the Institute of Computational Imaging, Beijing Information Science and Technology University, Beijing 102206, China (e-mail: 2024020607@bistu.edu.cn; sunshuchen1210@163.com; qiujun@bistu.edu.cn; changliuct@gmail.com).
Ligen Shi is with the College of Computer Science (College of Software), Inner Mongolia University, Hohhot 010021, China (e-mail: ligenshi0826@gmail.com).
Corresponding author: Chang Liu.
}}
\maketitle 
\thispagestyle{empty}

\begin{abstract}
Implicit neural representations provide a compact and continuous way to reconstruct dense light fields from sampled ray coordinates. However, fast light field reconstruction remains challenging because a light field is a high-dimensional signal with strong spatial--angular redundancy and structured disparity variations. Directly fitting 4D ray coordinates with a neural network often requires considerable optimization time to recover both view appearance and cross-view consistency. To address this issue, this paper proposes a fast implicit light field representation based on geometric decomposition and multi-resolution low-rank features. The proposed method decomposes a 4D light field into a horizontal disparity plane, a spatial texture plane, and a vertical disparity plane. Each plane is represented by a low-rank structure that combines a low-resolution 2D grid with the element-wise product of two high-resolution 1D line features at multiple resolution levels. The fused features are decoded by a lightweight multilayer perceptron to predict RGB values. Experiments on public light field datasets show that the proposed method achieves competitive reconstruction quality while providing a better trade-off among model parameters, training time, and inference efficiency.
\end{abstract}

\section{Introduction}
Light fields (LFs) record both the spatial positions and angular directions of scene rays, and have been used in free-viewpoint synthesis, 3D perception, digital refocusing, virtual reality, and autostereoscopic display \cite{levoy1996light,11293455}. These applications often require LF representations that support dense ray querying, compact storage, and efficient rendering. However, an LF is a high-dimensional signal with spatial--angular redundancy, and its large data volume remains a major challenge for real-time processing, transmission, and storage \cite{11068206}. Directly storing dense ray samples or fitting the full 4D signal with a generic neural network can lead to storage and optimization costs. Therefore, an efficient continuous representation is needed for LF modeling, especially when per-scene optimization and rendering efficiency are considered.

Under the standard two-plane parameterization, a light ray is represented by four coordinates $(u,v,x,y)$, where $(u,v)$ denote angular coordinates and $(x,y)$ denote spatial coordinates \cite{levoy1996light}. In this parameterization, the projection of the same scene point across different viewpoints forms disparity-dependent linear structures in epipolar plane images (EPIs) \cite{bolles1987epipolar,wanner2014variational}. The slope of an EPI line reflects scene disparity, indicating that angular and spatial coordinates in an LF are not independent but coupled by disparity. This spatial--angular coupling should be considered when designing an efficient implicit representation for LFs.

Implicit neural representations (INRs) model visual signals as continuous coordinate-based functions. Representative methods include NeRF~\cite{mildenhall2021nerf} for neural radiance field representation, SIREN~\cite{sitzmann2020siren} for implicit signal representation with periodic activation functions, and Fourier feature mapping~\cite{tancik2020fourier} for high-frequency coordinate fitting. To improve neural representation efficiency, explicit feature grids, tensor decomposition, and multi-resolution encoding have also been studied. Instant-NGP~\cite{muller2022instant} uses multi-resolution hash encoding, TensoRF~\cite{chen2022tensorf} represents radiance fields through tensor decomposition, K-Planes~\cite{fridovich2023kplanes} decomposes high-dimensional neural fields into 2D feature planes, and GA-Planes~\cite{sivgin2024gaplanes} introduces a low-rank plane representation. These methods indicate that structured feature representations can reduce the optimization burden of neural fields.

Neural light field (NLF) representations extend INR-based modeling to continuous light field representation and view synthesis. SIGNET~\cite{feng2021signet} introduces Gegenbauer polynomial encoding for compact NLF representation; NeuLF~\cite{li2022neulf} represents an LF as a mapping from 4D ray coordinates to color values; Light Field Networks~\cite{sitzmann2021light} support single-evaluation neural scene rendering; and Ray-Space Embedding models NLFs through learned ray-space embeddings~\cite{attal2022learning}. In addition to ray-color modeling, neural disparity field~\cite{shi2025iterative} uses an INR to represent a continuous disparity field reconstructed from discrete LF observations. Recently, ray displacement fields~\cite{liu2026neural} have been used to decompose NLF reconstruction into radiance representation and geometric displacement modeling. However, many existing NLF methods for ray-color representation still rely on direct 4D coordinate fitting or generic ray embeddings. Since the coordinates $(u,v,x,y)$ are coupled by disparity, directly fitting the full 4D function may require more parameters and optimization iterations to recover both view appearance and multi-view consistency.

Although generic feature decompositions can improve neural representation efficiency, they are not specifically designed for the EPI geometry of LFs. In rectified two-plane LFs, horizontal disparity is mainly represented by the coupling between $(u,x)$, vertical disparity by the coupling between $(v,y)$, and intra-view texture by $(x,y)$. This structure can be used as a prior for designing efficient light field INR models.

Motivated by these observations, this paper proposes a fast implicit NLF representation based on geometric decomposition and multi-resolution low-rank features. Instead of directly fitting the 4D LF with a generic coordinate multilayer perceptron (MLP), the proposed method decomposes the LF into a horizontal disparity plane, a spatial texture plane, and a vertical disparity plane. Each plane is represented by a multi-resolution low-rank structure, where a low-resolution 2D grid is combined with the element-wise product of two high-resolution 1D line features. The fused features are decoded by a lightweight MLP to predict RGB values. This design uses the spatial--angular structure of LFs while maintaining the continuous querying ability of INR models.

The main contributions of this paper are threefold. First, we propose a geometric tri-plane decomposition for LF INR, where a 4D LF is represented by a horizontal disparity plane, a spatial texture plane, and a vertical disparity plane, thereby considering the spatial--angular coupling in rectified LFs. Second, we introduce a multi-resolution low-rank plane representation that combines low-resolution 2D plane grids with high-resolution 1D line-feature products to reduce the redundancy of dense 2D feature grids. Third, we build an efficient continuous LF representation framework and evaluate its trade-offs among reconstruction quality, model parameters, training time, and inference efficiency on public LF datasets.

\section{Method}\label{sec:method}
\begin{figure*}[!t]
    \centering
    \includegraphics[width=\textwidth]{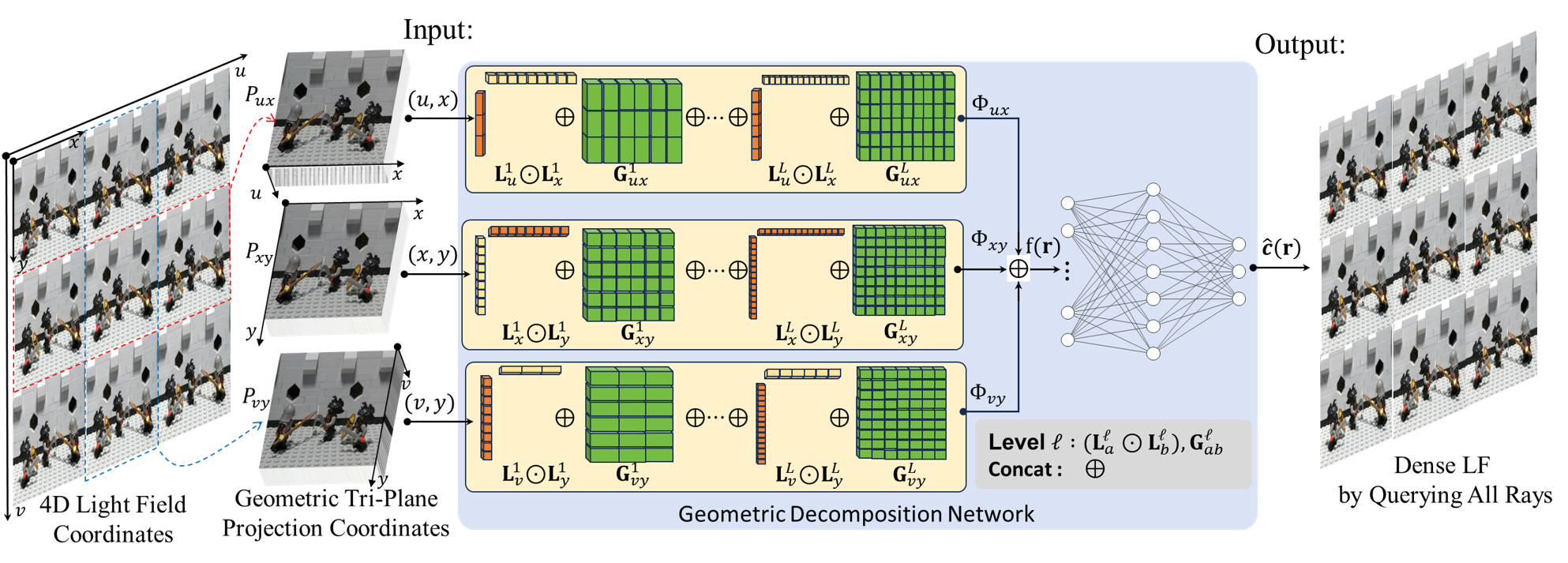}
    \caption{
    Overview of the proposed framework.
    A queried ray $\mathbf{r}=(u,v,x,y)$ is mapped onto three feature planes. Each plane uses a multi-resolution low-rank representation that combines a low-resolution 2D grid $\mathbf{G}_{ab}^{\ell}$ with the element-wise product of two high-resolution 1D line features $\mathbf{L}_{a}^{\ell}$ and $\mathbf{L}_{b}^{\ell}$.
    The concatenated features are decoded by a lightweight MLP to predict the ray color.
    }
    \label{fig:framework}
\end{figure*}

This section presents the proposed fast implicit LF representation. As shown in Fig.~\ref{fig:framework}, a queried ray $\mathbf{r}=(u,v,x,y)$ is mapped onto three geometry-related planes: the horizontal disparity plane ${P}_{ux}$, the spatial texture plane ${P}_{xy}$, and the vertical disparity plane ${P}_{vy}$. Each plane is modeled by a multi-resolution low-rank structure that combines a low-resolution 2D grid with high-resolution 1D line features. The extracted features are concatenated and decoded by a lightweight multilayer perceptron (MLP) to predict the RGB color.

\subsection{Problem Formulation}
\label{subsec:problem}

Under the two-plane parameterization, a light ray is represented as
\begin{equation}
    \mathbf{r}=(u,v,x,y),
\end{equation}
where $(u,v)$ and $(x,y)$ denote angular and spatial coordinates, respectively. Given an LF with observed ray-color samples, the training set is denoted as
\begin{equation}
    {D}
    =
    \{(\mathbf{r}_{i},\mathbf{c}_{i})\}_{i=1}^{N},
    \quad
    \mathbf{r}_{i}=(u_i,v_i,x_i,y_i),
    \quad
    \mathbf{c}_{i}\in\mathbb{R}^{3},
\end{equation}
where $N$ is the number of samples used for optimization. Implicit LF representation aims to learn a continuous mapping
\begin{equation}
    F_{\Theta}:\mathbf{r}\mapsto \mathbf{c},
\end{equation}
which predicts the RGB color for a ray coordinate within the LF domain.

Since LF coordinates are coupled by disparity, treating $(u,v,x,y)$ as an unstructured 4D input may be inefficient. We therefore decompose the LF into geometry-related 2D feature planes.

\subsection{Geometric Tri-Plane Decomposition}
\label{subsec:geometric_decomposition}

For rectified LFs, we define three 2D feature planes:
\begin{equation}
    S=\{(u,x),(x,y),(v,y)\}.
\end{equation}
Here, $(u,x)$, $(x,y)$, and $(v,y)$ correspond to the horizontal disparity plane ${P}_{ux}$, spatial texture plane ${P}_{xy}$, and vertical disparity plane ${P}_{vy}$, respectively.

For each $(a,b)\in{S}$, let $\Phi_{ab}(a,b)$ be the learnable feature function on plane ${P}_{ab}$. The structured feature of a queried ray is defined as
\begin{equation}
    \mathbf{f}(\mathbf{r})
    =
    \bigoplus_{(a,b)\in{S}}
    \Phi_{ab}(a,b),
\label{eq:structured_feature}
\end{equation}
where $\oplus$ denotes channel-wise concatenation.

\subsection{Multi-Resolution Low-Rank Plane Representation}
\label{subsec:low_rank_representation}

Each plane ${P}_{ab}$ is represented by $L$ resolution levels. At the $\ell$-th level, we use a low-resolution 2D grid $\mathbf{G}_{ab}^{\ell}$ and two high-resolution 1D line features $\mathbf{L}_{a}^{\ell}$ and $\mathbf{L}_{b}^{\ell}$. The grid feature is obtained by
\begin{equation}
    \mathbf{g}_{ab}^{\ell}(a,b)
    =
    \mathcal{\psi}_{2D}
    \left(
        \mathbf{G}_{ab}^{\ell},a,b
    \right),
\label{eq:plane_grid_feature}
\end{equation}
where $\mathcal{\psi}_{2D}(\cdot)$ denotes bilinear interpolation.

The line-product feature is defined as
\begin{equation}
\begin{aligned}
    \mathbf{q}_{ab}^{\ell}(a,b)
    =
    \mathcal{\psi}_{1D}
    \left(
        \mathbf{L}_{a}^{\ell},a
    \right)
    \odot
    \mathcal{\psi}_{1D}
    \left(
        \mathbf{L}_{b}^{\ell},b
    \right),
\end{aligned}
\label{eq:line_product_feature}
\end{equation}
where $\mathcal{\psi}_{1D}(\cdot)$ denotes linear interpolation and $\odot$ denotes element-wise multiplication.

The final feature on plane ${P}_{ab}$ is
\begin{equation}
    \Phi_{ab}(a,b)
    =
    \bigoplus_{\ell=1}^{L}
    \left[
        \mathbf{g}_{ab}^{\ell}(a,b)
        \oplus
        \mathbf{q}_{ab}^{\ell}(a,b)
    \right].
\label{eq:plane_feature}
\end{equation}
This representation combines coarse 2D plane features with separable 1D line-product features, reducing parameter growth compared with dense high-resolution 2D grids.

\subsection{Color Decoding and Training Objective}
\label{subsec:decoder_loss}
Given the structured feature $\mathbf{f}(\mathbf{r})$, a lightweight decoder $D_{\theta}$ predicts the ray color:
\begin{equation}
    \hat{\mathbf{c}}(\mathbf{r})
    =
    D_{\theta}
    \left(
        \mathbf{f}(\mathbf{r})
    \right),
\label{eq:complete_model}
\end{equation}
where the overall LF representation is denoted as
$F_{\Theta}(\mathbf{r})=D_{\theta}(\mathbf{f}(\mathbf{r}))$.

For a mini-batch $B\subset D$, the loss is
\begin{equation}
    \mathcal{L}_{\mathrm{r}}
    =
    \frac{1}{|{B}|}
    \sum_{(\mathbf{r}_{i},\mathbf{c}_{i})\in{B}}
    \left\|
        F_{\Theta}(\mathbf{r}_{i})
        -
        \mathbf{c}_{i}
    \right\|_{2}^{2}.
\label{eq:rgb_loss}
\end{equation}
The model parameters are optimized by
\begin{equation}
    \Theta^{*}
    =
    \arg\min_{\Theta}
    \mathcal{L}_{\mathrm{r}},
\label{eq:optimization}
\end{equation}
where $\Theta$ includes the plane grids, line features, and decoder parameters. After optimization, Eq.~\eqref{eq:complete_model} can be used to query continuous ray coordinates within the LF domain.

\section{Experiments}
\label{sec:experiments}

This section evaluates the proposed method on public LF datasets from three aspects: reconstruction quality, representation efficiency, and the effectiveness of the proposed structural design.

\subsection{Experimental Setup}
\label{subsec:experimental_setup}
We evaluate the proposed method on three public LF datasets: EPFL (10 scenes), INRIA Lytro (5 scenes), and Stanford Gantry (2 scenes). These real and controlled-capture datasets have been used in the NTIRE 2025 LF Image Super-Resolution benchmark~\cite{wang2025ntire}, and include different textures, depth variations, occlusions, and view-dependent effects. Reconstruction quality is measured by PSNR, SSIM, and LPIPS, while efficiency is evaluated by model parameters, training time, and inference time.

All methods are evaluated under the same $9\times9$ LF setting on an NVIDIA GeForce RTX 4090 GPU with 24 GB memory. The implementation is based on Python 3.12 and PyTorch 2.8.0 with CUDA mixed-precision training. Our model uses three geometry-related planes, $P_{ux}$, $P_{xy}$, and $P_{vy}$, each containing a 2D grid branch and a 1D line-product branch. For the 2D grid branch, the spatial plane $P_{xy}$ uses 8 square-grid levels ranging from $16\times16$ to $256\times256$. For the two disparity planes $P_{ux}$ and $P_{vy}$, the angular resolution ranges from 3 to 9 and is paired with the spatial resolution ranging from 16 to 128 at each level, forming multi-resolution 2D grids from $3\times16$ to $9\times128$. In the 1D line-product branch, the spatial line resolutions range from 64 to 512, while the angular line resolutions follow the corresponding angular levels from 3 to 9. Each feature level has 8 channels, each plane outputs a 128-dimensional feature, and the concatenated tri-plane feature has 384 dimensions. The decoder is a ReLU MLP with hidden width 256 and a linear RGB output layer. Following the optimization practice in high-fidelity INR reconstruction~\cite{mcginnis2025optimizing}, we train the model for 300 epochs using Muon for the hidden-layer weights of the MLP decoder and Adam for the remaining parameters, with learning rates of 0.02 and $1\times10^{-2}$, respectively, together with a cosine annealing scheduler.

We compare the proposed method with three representative NLF methods, including SIGNET, RSEN, and RDF. For all baselines, we follow the settings and default hyperparameters reported in the original papers.

\begin{table}[!t]
\centering
\footnotesize
\setlength{\tabcolsep}{4pt}
\renewcommand{\arraystretch}{0.95}
\caption{Quantitative comparison on public LF datasets. Best results are marked in bold. The average is computed over datasets.}
\label{tab:quantitative_results}
\begin{tabular}{llccc}
\toprule
Dataset & Method & PSNR (dB)$\uparrow$  & SSIM$\uparrow$ & LPIPS$\downarrow$ \\
\midrule
\multirow{4}{*}{EPFL}
 & SIGNET & 33.96$\pm$2.67 & 0.93$\pm$0.04 & 0.09$\pm$0.05 \\
 & RSEN   & 30.10$\pm$1.64 & 0.87$\pm$0.05 & 0.19$\pm$0.06 \\
 & RDF    & 31.85$\pm$1.73 & \textbf{0.94$\pm$0.02} & 0.07$\pm$0.05 \\
 & Ours   & \textbf{35.45$\pm$2.55} & 0.93$\pm$0.03 & \textbf{0.02$\pm$0.02} \\
\midrule
\multirow{4}{*}{INRIA}
 & SIGNET & 35.43$\pm$3.07 & 0.91$\pm$0.06 & 0.12$\pm$0.05 \\
 & RSEN   & 30.99$\pm$1.81 & 0.85$\pm$0.09 & 0.24$\pm$0.07 \\
 & RDF    & 30.40$\pm$3.63 & 0.90$\pm$0.04 & 0.10$\pm$0.05 \\
 & Ours   & \textbf{37.61$\pm$3.61} & \textbf{0.92$\pm$0.05} & \textbf{0.04$\pm$0.03} \\
\midrule
\multirow{4}{*}{Stanford}
 & SIGNET & \textbf{39.37$\pm$2.68} & 0.97$\pm$0.07 & 0.04$\pm$0.01 \\
 & RSEN   & 36.76$\pm$0.77 & 0.97$\pm$0.04 & 0.07$\pm$0.04 \\
 & RDF    & 29.36$\pm$4.09 & 0.95$\pm$0.03 & 0.03$\pm$0.02 \\
 & Ours   & 37.30$\pm$5.22 & \textbf{0.97$\pm$0.01} & \textbf{0.01$\pm$0.06} \\
\midrule
\multirow{4}{*}{Average}
 & SIGNET & 36.25$\pm$2.80 & \textbf{0.94$\pm$0.03} & 0.08$\pm$0.04 \\
 & RSEN   & 32.63$\pm$3.69 & 0.90$\pm$0.06 & 0.16$\pm$0.09 \\
 & RDF    & 30.54$\pm$1.25 & 0.93$\pm$0.03 & 0.07$\pm$0.04 \\
 & Ours   & \textbf{36.79$\pm$1.17} & \textbf{0.94$\pm$0.03} & \textbf{0.02$\pm$0.02} \\
\bottomrule
\end{tabular}
\end{table}

\begin{figure*}[!t]
    \centering
    \includegraphics[width=1\textwidth]{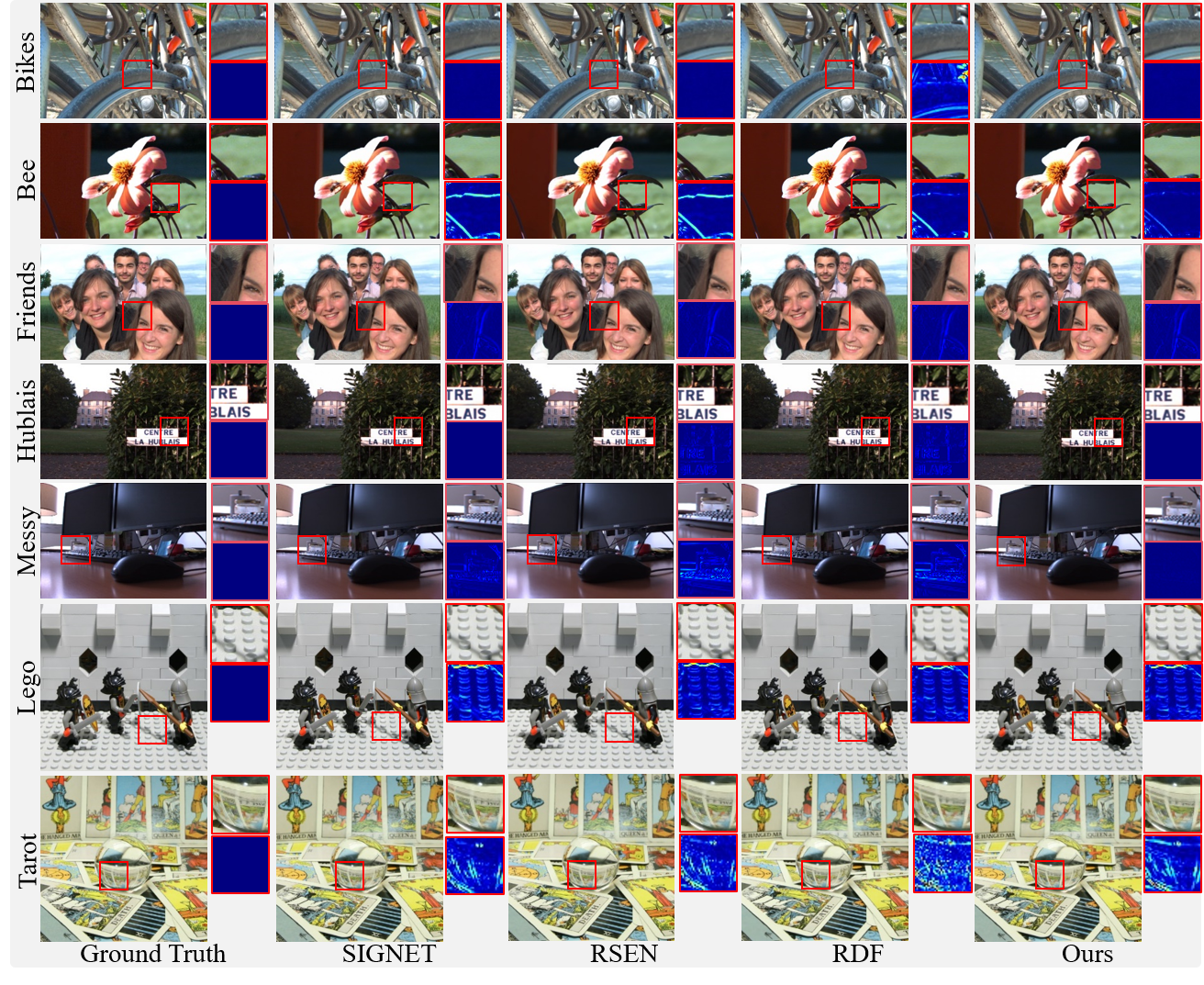}
    \caption{Qualitative visual comparison of different methods on representative images selected from the EPFL, INRIA Lytro, and Stanford Gantry datasets, including the challenging Tarot scene. The results are shown for the center views, including reconstructed images and corresponding error maps.}
    \label{fig:qualitative_results}
\end{figure*}

\subsection{Quantitative and Qualitative Results}
\label{subsec:quantitative_qualitative_results}

Table~\ref{tab:quantitative_results} reports quantitative results on EPFL, INRIA Lytro, and Stanford Gantry. Our method achieves the best macro-average PSNR and LPIPS, with SSIM comparable to the strongest baseline. It obtains the best PSNR and LPIPS on EPFL and INRIA Lytro, while achieving the lowest LPIPS on Stanford Gantry, showing a favorable balance between reconstruction quality and perceptual fidelity.

Fig.~\ref{fig:qualitative_results} shows representative reconstruction results and error maps using pixel-wise absolute differences. Compared with SIGNET, RSEN, and RDF, our method generally preserves finer structures and produces lower error responses around textures, edges, and occlusion boundaries. However, in the challenging Tarot scene with specular reflection and transparent glass, local blurring and ghosting artifacts can still be observed. This suggests that highly view-dependent non-Lambertian effects may violate the low-rank assumption of the proposed representation and remain a limitation of our method.

\subsection{Efficiency Analysis}\label{subsec:efficiency_analysis}

\begin{table}[!t]
\centering
\caption{Efficiency comparison of different methods on the INRIA Lytro dataset.}
\label{tab:efficiency_results}
\footnotesize
\setlength{\tabcolsep}{2.5pt}
\renewcommand{\arraystretch}{0.95}
\begin{tabular}{@{}lccc@{}}
\toprule
Method & Params. (M) & Train. (min) & Infer. (s)\\
\midrule
SIGNET & 2.37 & 360 & 20 \\
RSEN   & 1.40 & 180 & 9  \\
RDF    & 0.66 & 72  & 12 \\
Ours   & 1.55 & 21  & 8  \\
\bottomrule
\end{tabular}
\end{table}

\begin{table}[!t]
\centering
\caption{Training time required to reach the same target PSNR on the INRIA Lytro dataset.}
\label{tab:time_to_psnr}
\footnotesize
\setlength{\tabcolsep}{4.5pt}
\renewcommand{\arraystretch}{1.08}
\resizebox{\columnwidth}{!}{
\begin{tabular}{lccccc}
\toprule
\multirow{2}{*}{Scene} 
& \multirow{2}{*}{Target PSNR (dB)} 
& \multicolumn{4}{c}{Train.(min)}  \\
\cmidrule(lr){3-6}
& 
& SIGNET 
& RSEN 
& RDF 
& Ours \\
\midrule
Bee       & 25.33 & 11  & 29  & 75 & \textbf{0.8}  \\
Building  & 29.93 & 37  & 168 & 70 & \textbf{2.0}  \\
Hublai    & 29.17 & 45  & 155 & 73 & \textbf{1.5} \\
MessyDesk & 33.44 & 72  & 173 & 62 & \textbf{2.0} \\
Sculpture & 32.19 & 60  & 180 & 68 & \textbf{2.0}  \\
\midrule
Average   & 30.01 & 45.0 & 141.0 & 69.6 & \textbf{1.7}  \\
\bottomrule
\end{tabular}
}
\end{table}

Since one of the main objectives of the proposed method is efficient light field reconstruction, Table~\ref{tab:efficiency_results} compares the model parameters, average convergence time, and inference time of different methods under their default configurations. Our method achieves the shortest average convergence time and inference time with a moderate number of parameters.  To further ensure a fair comparison, Table~\ref{tab:time_to_psnr} reports the training time required to reach the same target PSNR on five INRIA Lytro scenes. The results show that our method remains significantly faster than SIGNET, RSEN, and RDF under the same reconstruction quality. This advantage comes from the proposed geometric tri-plane decomposition and multi-resolution low-rank features, which reduce the optimization burden of direct 4D coordinate fitting and avoid explicitly optimizing an additional displacement field.

\begin{table}[!t]
\centering
\footnotesize
\setlength{\tabcolsep}{3pt}
\renewcommand{\arraystretch}{0.95}
\caption{Ablation study results on the INRIA Lytro dataset.}
\label{tab:ablation_results}
\begin{tabular}{lccc}
\toprule
Setting & PSNR (dB)$\uparrow$ & SSIM$\uparrow$ & LPIPS$\downarrow$ \\
\midrule
w/o multi-res. & 34.67$\pm$3.21 & 0.90$\pm$0.06 & 0.05$\pm$0.03 \\
w/o line prod. &35.46$\pm$3.21 & 0.91$\pm$0.05 & 0.04$\pm$0.02 \\
w/o 2D grid.  & 33.60$\pm$3.26 & 0.87$\pm$0.07 & 0.09$\pm$0.05 \\
Full model & \textbf{37.61$\pm$3.61} & \textbf{0.92$\pm$0.05} & \textbf{0.04$\pm$0.03} \\
\bottomrule
\end{tabular}
\end{table}

\begin{table}[!t]
\centering
\footnotesize
\setlength{\tabcolsep}{4pt}
\renewcommand{\arraystretch}{0.95}
\caption{Impact of multi-resolution levels on the Stanford Gantry dataset.}
\label{tab:level_sensitivity}
\begin{tabular}{lcccc}
\toprule
Levels & PSNR (dB)$\uparrow$ & SSIM$\uparrow$ & LPIPS$\downarrow$ & Train. (min) \\
\midrule
4 levels & 36.10$\pm$5.61 & 0.96$\pm$0.02 & 0.02$\pm$0.01 & 13 \\
6 levels & 36.91$\pm$5.36 & 0.96$\pm$0.01 & 0.02$\pm$0.01 & 17 \\
8 levels & \textbf{37.30$\pm$5.22} & \textbf{0.97$\pm$0.01} & \textbf{0.01$\pm$0.01} & 21 \\
\bottomrule
\end{tabular}
\end{table}

\subsection{Ablation Study}
\label{subsec:ablation_study}
We conduct ablation experiments on the INRIA Lytro dataset (5 scenes) to evaluate the contribution of the main components, including the multi-resolution design, the 1D line-feature product, and the 2D plane grid. As shown in Table~\ref{tab:ablation_results}, the full model achieves the best PSNR and SSIM, while maintaining the best or comparable LPIPS. In addition, we analyze the influence of the number of resolution levels on the Stanford Gantry dataset (2 scenes), as shown in Table~\ref{tab:level_sensitivity}. Increasing the number of levels improves reconstruction quality by capturing richer multi-scale details. Although more levels increase training time, the additional cost is acceptable compared with the quality gain. Therefore, we use the 8-level setting as the default configuration.

\section{Conclusion}\label{sec:conclusion}
This paper presented a fast implicit LF representation based on geometric decomposition and multi-resolution low-rank features. The proposed method decomposes a 4D LF into three geometry-related planes and represents each plane using low-resolution 2D grids and high-resolution 1D line-feature products. This design reduces the redundancy of dense 2D feature grids while preserving structured LF information. Experiments on public datasets show that the proposed method provides a favorable trade-off among reconstruction quality, model parameters, training time, and inference efficiency. Future work will explore sparse-view reconstruction, dynamic LFs, and real-time rendering.

\section*{Acknowledgment}
This work was supported by National Natural Science Foundation of China (under Grant Nos. 62171044, 61931003).
\medskip
\bibliography{ref} 
\bibliographystyle{unsrt}
\end{document}